\documentclass[runningheads]{llncs}
\usepackage{graphicx}
\usepackage{amsmath,amssymb} % define this before the line numbering.
\usepackage{color}

\usepackage{url}
 
\usepackage{bm}
\usepackage{algorithm}
\usepackage{algpseudocode}
\usepackage{multirow}
\usepackage[tight,footnotesize]{subfigure}
\usepackage{caption}
\usepackage{hhline}
\usepackage{wrapfig}

\algnewcommand\algorithmicforeach{\textbf{for each}}
\algdef{S}[FOR]{ForEach}[1]{\algorithmicforeach\ #1\ \algorithmicdo}

\usepackage[whole]{bxcjkjatype}	% 日本語メモ用

\newcommand{\Tref}[1]{Table~\ref{#1}}
\newcommand{\eref}[1]{Eq.~(\ref{#1})}

\newcommand{\fref}[1]{Fig.~\ref{#1}}
\newcommand{\Fref}[1]{Figure~\ref{#1}}
\newcommand{\sref}[1]{Sec.~\ref{#1}}

\newcommand{\argmin}{\mathop{\rm argmin}\limits}

\newcommand{\vi}{\mathbf{i}}
\newcommand{\vx}{\mathbf{x}}

\newcommand{\vp}{\mathbf{p}}

\newcommand{\vt}{\mathbf{t}}
\newcommand{\band}{\mathcal{B}} % \fo{Epipolar band: set of lines (`b'and)}
\newcommand{\region}{\mathcal{A}}   % \fo{Instance region: set of pixels (`a'rea); avoiding to use R (`r'egion) not to confuse with \mathbb{R}. }
\newcommand{\epimap}{\mathcal{M}}   % \fo{Epipolar map: set of pixels (`m'ap). }

\newcommand{\ie}{\textit{i}.\textit{e}.}
\newcommand{\etal}{\textit{et~al}.}
\newcommand{\eg}{\textit{e}.\textit{g}.}

\begin{document}
\pagestyle{headings}
\mainmatter

\def\ACCV20SubNumber{301}  % Insert your submission number here

%===========================================================
\title{Descriptor-Free Multi-View Region Matching for Instance-Wise 3D Reconstruction}
\titlerunning{Descriptor-Free Multi-View Region Matching}

\author{Takuma Doi\inst{1} \and
Fumio Okura\inst{1,2}\orcidID{0000-0001-7595-1300} \and
Toshiki Nagahara\inst{1} \and
Yasuyuki Matsushita\inst{1}\orcidID{0000-0002-1935-4752}\and
Yasushi Yagi\inst{1}}
\authorrunning{T. Doi et al.}

\institute{Graduate School of Information Science and Technology,\\Osaka University, Osaka, Japan\\
\email{\{doi,t-nagahara,yagi\}@am.sanken.osaka-u.ac.jp}\\ \email{\{okura,yasumat\}@ist.osaka-u.ac.jp} \and
Japan Science and Technology Agency, Saitama, Japan}

\maketitle

%===========================================================
\begin{abstract}
This paper proposes a multi-view extension of instance segmentation without relying on texture or shape descriptor matching.
Multi-view instance segmentation becomes challenging for scenes with repetitive textures and shapes, \eg, plant leaves, due to the difficulty of multi-view matching using texture or shape descriptors. 
To this end, we propose a multi-view region matching method based on epipolar geometry, which does not rely on any feature descriptors. We further show that the epipolar region matching can be easily integrated into instance segmentation and effective for instance-wise 3D reconstruction. 
Experiments demonstrate the improved accuracy of multi-view instance matching and the 3D reconstruction compared to the baseline methods.
\end{abstract}

%===========================================================
\section{Introduction}

\begin{figure}[t]
	\centering
	\includegraphics[width=\linewidth]{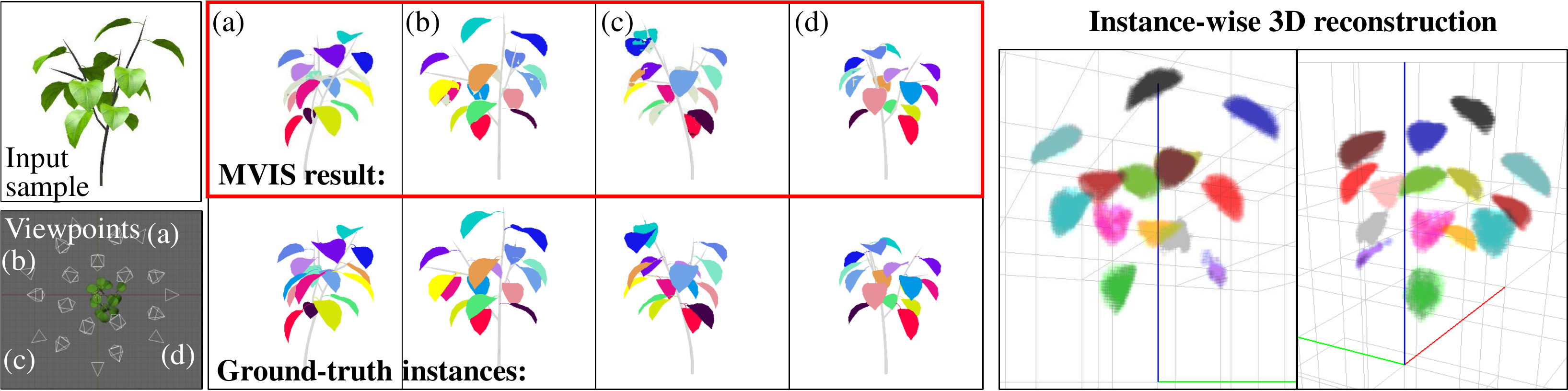}  
    \captionof{figure}{An example of MVIS and instance-wise 3D reconstruction. The same color indicates the corresponding instances across different views. We propose a multi-view instance matching without relying on texture or shape descriptors.}
	\label{fig:teaser}
\end{figure}

One of the important applications of computer vision is to analyze the shape and structure of natural objects, \ie, quantifying the size, number, and shape of plant leaves. In such applications, \emph{instance segmentation} plays an important role because of the need for accurately determining individual object instances.
A single-view instance segmentation has gained attention, and there have been a series of previous works~\cite{maskrcnn,ren2017end,xiong2019upsnet} to address the problem. 
However, for scenes suffering from severe occlusions, it is needed to perform instance segmentation in a \emph{multi-view} setting as shown in \fref{fig:teaser}. 
We call it a \emph{multi-view instance segmentation (MVIS)} problem.

An extension of single-view instance segmentation to the multi-view setting is a challenging task, especially when the individual instances are look-alike, for example, plant leaves.
Although external markers or background objects, \eg, plant pots, can be a guide for obtaining sparse correspondences to determine camera poses, we cannot expect dense point correspondences on similar instances, \eg, leaves, across views.
This setting makes it difficult to directly use existing approaches, such as 3D instance segmentation~\cite{hou20193d} and multi-view semantic segmentation~\cite{kowdle2012multiple,djelouah2013multi,mustafa2017semantically}, which heavily rely on 3D shape information.

To overcome the problem, we propose an MVIS method based on region matching that does not rely on dense correspondences, which can be used for matching objects across views without distinctive textures or shape but with similar appearances.
The key idea of our method is to use the epipolar constraint to determine the multi-view region correspondences of segmented instances that are computed at each view. 
By treating each instance in each view as a node in a graph, and by assigning edge weights based on the degree of intersection of epipolar lines and instance regions, we cast the problem of multi-view instance segmentation to a graph clustering problem. 
We call the approach the \emph{epipolar region matching}. We show that it enables reliably establishing correspondences of view-wise segmented instances across multiple views. Further, our epipolar region matching can be easily integrated in modern instance segmentation methods~\cite{maskrcnn,xiong2019upsnet,voigtlaender2019mots}. 
Our method can be combined with any region segmentation methods; while given a set of regions, our region matching method does not rely on any descriptors.
The estimated instances with multi-view correspondences can be used for the \emph{instance-wise 3D reconstruction} via existing 3D reconstruction methods, yielding superior reconstruction results.

Experimental results show that the proposed method yields accurate multi-view instance correspondence compared to an instance matching method based on a traditional multi-view stereo (MVS) and a point-based matching. We also demonstrate the effectiveness of MVIS for instance-wise 3D reconstruction.

\paragraph{Contributions.} 
We propose a multi-view region matching method without using texture and shape descriptors, as a key technical component of MVIS for texture/shape-repetitive scenes.
The matching method can be put together in instance segmentation and 3D reconstruction methods for instance-wise 3D reconstruction.

\section{Related Works}
Our work is closely related to multi-view correspondence matching, as well as the multi-view or 3D extensions of semantic/instance segmentation. This section discusses the related works in these subject areas.

\subsubsection{Multi-view correspondence matching.} 
Multi-view correspondence matching is a fundamental problem in multi-view image analysis, such as structure from motion (SfM)~\cite{schoenberger2016sfm} and multi-view stereo (MVS)~\cite{schoenberger2016mvs}.
As a veteran but still frequently used approach, keypoint detection and matching (\eg,~\cite{lowe2004distinctive,alcantarilla2011fast,rublee2011orb}) is used to obtain sparse correspondences; while dense correspondence matching often involves a patch~\cite{bleyer2011patchmatch} and plane-based~\cite{gallup2007real} matching.

%We focus on the matching problem in scenes without distinctive shape and texture features.
For target scenes containing partly texture-less objects, various approaches have been studied, such as a belief propagation~\cite{sun2003stereo,furukawa2004structure} for MVS. 
More recently, deep-learning-based approaches to correspondence matching for partially texture-less scenes, which often assume the smoothness of the target scene, have been studied~\cite{zhang2019learning,romanoni2019tapa}.
Correspondence matching for objects in which surface is fully texture-less, or with repetitive textures is notably challenging because it becomes difficult to find dense correspondences.
For string-like objects (\eg, hairs), which have the matching ambiguity along the strings or lines, an MVS method using line-shaped patches has been proposed~\cite{nam2019strand}. 
Regarding the sparse point matching, Dellaert~\etal~\cite{dellaert2000structure} presented a method for the SfM problem without using texture features, which calculates the camera pose and sparse point correspondences based only on the geometric relationship of keypoints.

Shape-based matching~\cite{belongie2002shape,berg2005shape,mikolajczyk2005comparison} such as using the Fourier shape descriptors~\cite{bartolini2005warp} or shape contexts~\cite{belongie2002shape}, is another line of the region matching. These approaches encode the region shape information, \eg, instance shape, into descriptors.
In our cases, since the target scene is composed of objects with similar 3D shapes (\eg, plant leaves), it is not realistic to use shape-based matching.
Although it is different from the context of instance segmentation, region-based information~\cite{matas2004robust}, such as the statistics of textures, is a useful cue for the correspondence matching.
For example, superpixel stereo~\cite{li2016pmsc,mivcuvsik2010multi} or segment-based stereo~\cite{klaus2006segment} were developed to increase the robustness of correspondence matching.
Segmentation of the same object among multiple views can be categorized in multi-view co-segmentation~\cite{kowdle2012multiple,djelouah2013multi,mustafa2017semantically}, which also uses the texture features.% as well as geometric constraint.

Unlike the previous methods, we achieve the descriptor-free multi-view region matching using the epipolar constraint.
The epipolar matching is a traditional yet important problem. As far as we are aware, \emph{searching regions without (texture or shape) descriptor matching} remains unsolved.

\vspace{-1.5em}
\subsubsection{Instance segmentation and its multi-view extension.}
%The proposed approach highly relies on the development of recent instance segmentation methods.
While various implementations of instance segmentation including one-stage methods~\cite{kulikov2020instance} are developed, two-stage frameworks that combine object detection and mask generation are often used due to the advantage of performance and simplicity. A major implementation of two-stage instance segmentation is Mask R-CNN~\cite{maskrcnn} based on Faster R-CNN~\cite{ren2015faster}, which computes the region proposals of the target objects followed by the mask generation for selected proposals.

Video instance segmentation is a multi-image extension of instance segmentation.
These approaches perform the instance tracking, as well as the segmentation~\cite{milan2015joint,jun2017cdts,ovsep2018track,sharma2018beyond}.
An early attempt of this approach~\cite{seguin2016instance} used superpixels for instance tracking. 
A recent study configures the problem as multi-object tracking and segmentation (MOTS), and provides a Mask-R-CNN-based implementation~\cite{voigtlaender2019mots}.
PanopticFusion~\cite{narita2019panopticfusion} aggregates multi-view instances in a 3D volumetric space, while it inputs a video sequence that enables the instance tracking.
Video instance tracking is applied for plant image analysis, such as the 3D reconstruction of grape clusters~\cite{scholer2015automated}. 
This batch of approaches assumes the time-series image sequences, which allows the object tracking with smaller displacement between the consequent frames.
We rather focus on the multi-view setting, where the images are captured with wider baselines between the viewpoints.

\vspace{-1.5em}
\subsubsection{3D semantic/instance segmentation.}
Somewhat related to MVIS, 3D semantic/instance segmentation is studied.
Approaches for 3D segmentation often input 3D shapes of target objects, such as point cloud~\cite{qi2017pointnet,wang2019multi}, volumetric space~\cite{lahoud20193d}, and RGB-D images~\cite{hou20193d,dai20183dmv}.
Semantic segmentation on multi-view RGB images can be used for 3D reconstruction via the aggregation on a 3D space, which are called semantic 3D reconstruction~\cite{hane2013joint,savinov2015discrete,savinov2016semantic}.
These methods do not distinguish the objects in a same category in principle.% because they do not solve the instance correspondence problem.

Nassar~\etal~\cite{nassar19} proposes an instance warping for multi-view instance correspondence matching using the 3D shape of target scenes.
Instance segmentation on 3D point clouds are also studied~\cite{wang2018sgpn,yi2019gspn,engelmann20203d}, which needs the input of high-quality 3D point clouds.
Unlike these approaches, we focus on the instance matching of objects where the 3D reconstruction is challenging; \eg, plants due to thin shape, repetitive textures, and heavy occlusions.
As an application-specific study, a plant modeling method proposed by Quan~\etal~\cite{quan06} uses the combination of 2D segmentation and 3D point cloud clustering for 3D leaf modeling. Because their method assumes a good-quality 3D point cloud as input, it is difficult for the reconstruction of plants with texture-less leaves.

\section{Multi-view instance segmentation (MVIS)}

Our goal is to find multi-view instance correspondences from the multi-view images of the target. We assume that the camera poses and intrinsics are known, \eg, by an SfM~\cite{schoenberger2016sfm} using the sparse correspondences obtained from the scene background. We here assume that instance segmentation in each view is available for now, while the instance correspondence \emph{across views} is unknown. 

\subsection{Epipolar region matching}
\label{sec:epipolar}
\begin{figure}[t]
\begin{center}
    \includegraphics[width=\linewidth]{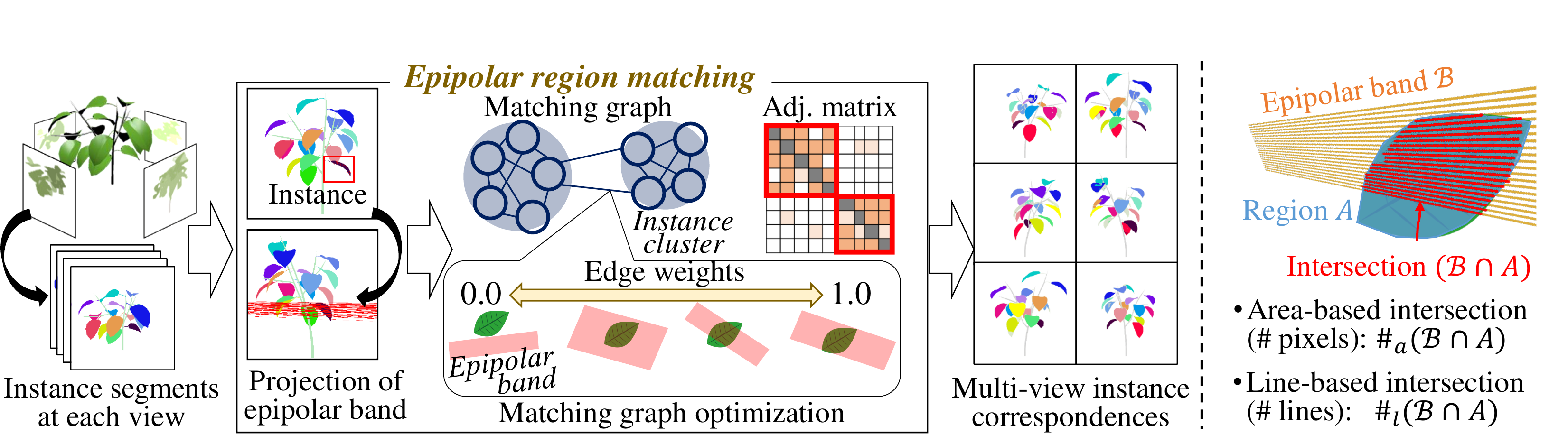}
\end{center}
   \caption{Overview of epipolar region matching which does not rely on the feature descriptors. The proposed method creates a matching graph based on the epipolar constraint, where the edge weights are defined as the degree of intersection between the instances and epipolar bands (\ie, projections of instances in other views). Instance correspondences are extracted by a graph clustering method.} 
\label{fig:overview}
\label{fig:iou}
\end{figure}

Given a set of instances in each image and camera pose information, we establish instance correspondences across multi-view images. 
We approach this problem by epipolar geometry; namely, finding the correspondences via epipolar lines drawn on other views. Since multiple instances appear on the same epipolar line (see \fref{fig:overview}), we cast the matching problem to a \emph{graph clustering} problem.

We create an undirected edge-weighted graph \hbox{$G=(V,E,w)$}, where $V$ is the node set that consists of instances appeared in all views, \hbox{$E \subseteq V \times V$} is the edge set, and \hbox{$w: E \rightarrow \mathbb{R}_+$} is the weight function defining the edge weights. The number of nodes $|V|$ corresponds to the total number of instances that appear in all the images.

To define the edge weights, we compute a set of epipolar lines from densely-sampled points on each instance.
Here we define the $i$-th view image as $I_i$, $m$-th instance segment in the $i$-th image as $v_{i,m} \in V$ and its region (set of pixels) in the image coordinates as $\region_{i,m} \subset \mathbb{R}^2$. 
Let $\mathbf{F}_{ij}$ be  a fundamental matrix between $i$-th and $j$-th views. Then an image point $\mathbf{p}$ in instance segment $\region_{i,m}$, \ie, $\mathbf{p} \in \region_{i,m}$, forms an epipolar line $\mathbf{l}_{j}(\mathbf{p})$ in the $j$-th view image as
\begin{eqnarray}
\mathbf{l}_{j}(\mathbf{p}) = \mathbf{F}_{ij} \tilde{\mathbf{p}},
\end{eqnarray}
where $\tilde{\mathbf{p}}$ is the homogeneous representation of $\mathbf{p}$. From all points in region $\region_{i,m}$, we have a set of epipolar lines forming a pencil of epipolar lines passing through the epipole. In what follows, we call the pencil an \emph{epipolar band} $\band$ in this paper. The epipolar band $\band$ of region $\region_{i,m}$ on the $j$-th image is thus defined as
\begin{eqnarray}
\band(\region_{i,m}, I_j) = \{\mathbf{F}_{ij} \tilde{\mathbf{p}}\}, \quad \mathbf{p} \in \region_{i,m}.
\end{eqnarray}

We define the edge weights $w$ in the graph $G$ as the degree of intersection of an epipolar band $\band$ and instance regions $\region$ in analogous to the intersection-over-union (IoU) computation. 
While original IoU is defined between two areas, the extension to a similarity measure between the epipolar band and a region is not straightforward.
To evaluate the degree of intersection between the epipolar band and instance region, we use two measures: The area of intersection and the number of epipolar lines in the epipolar band passing through the instance region, as illustrated in the right side of \fref{fig:iou}.
In this manner, the edge weight $w$ between nodes $v_{i,m}$ and $v_{j,n}$ can be obtained as
\begin{eqnarray}
    \!\! w(v_{i,m},v_{j,n}) =
    \frac{\#_a(\band(\region_{i,m},I_j) \cap \region_{j,n})}{ \#_a \region_{j,n}} \cdot  \frac{ \#_l (\band(\region_{i,m},I_j) \cap \region_{j,n})}{ \#_l \band(\region_{i,m},I_j)},
\end{eqnarray}
where the function $\#_a$ counts the number of pixels belonging to the region, while $\#_l$ counts the number of epipolar lines passing through the area. 
During the computation of $\#_a(\band(\region_{i,m},I_j)\cap \region_{j,n})$, we draw epipolar lines with the same thickness (two pixels was used in our experiment).
As with many other similarity measures, $w(v_{i,m},v_{j,n})$ takes the range $[0,1]$, where it becomes one if all epipolar lines in the epipolar band passes through the instance area and a whole part of the area is filled with epipolar lines (see \fref{fig:overview}).

Once the edge weights $\{w\}$ are defined, we form an adjacency matrix $\mathbf{W}$ of $|V| \times |V|$ elements from the graph $G$ and perform a graph clustering using symmetric non-negative matrix factorization (SymNMF)~\cite{kuang2012symmetric}, as $\mathbf{W} \approx \mathbf{H}\mathbf{H}^T$. 
After the factorization, the largest element in each row of $\mathbf{H}$ indicates the \emph{cluster ID}, where each cluster forms a set of corresponding instances across multi-view. Hereafter, we call the cluster an \emph{instance cluster}. 

\subsection{Application: Epipolar region matching for region proposals}
As an application of the proposed method, we here describe the integration of multi-view region matching method with modern instance segmentation.
While our region matching method can be used with any instance segmentation methods, we can optionally utilize the correspondences across multi-view \emph{region proposals} to retain partly occluded instances.
Most instance segmentation methods~\cite{maskrcnn,xiong2019upsnet,voigtlaender2019mots} rely on region proposals~\cite{ren2015faster}, which performs region unification by the non-maximum suppression (NMS)~\cite{girshick2015deformable} to merge overlapping region proposals. As a result, region proposals for partially occluded instances are often suppressed. 
To this end, we implemented an NMS process considering the multi-view region correspondences. We found our implementation recovered partially occluded instances, while it did not significantly affect the overall instance segmentation accuracy (see the supplementary material for detailed discussions).

\begin{figure}[t]
\begin{center}
    \includegraphics[width=\linewidth]{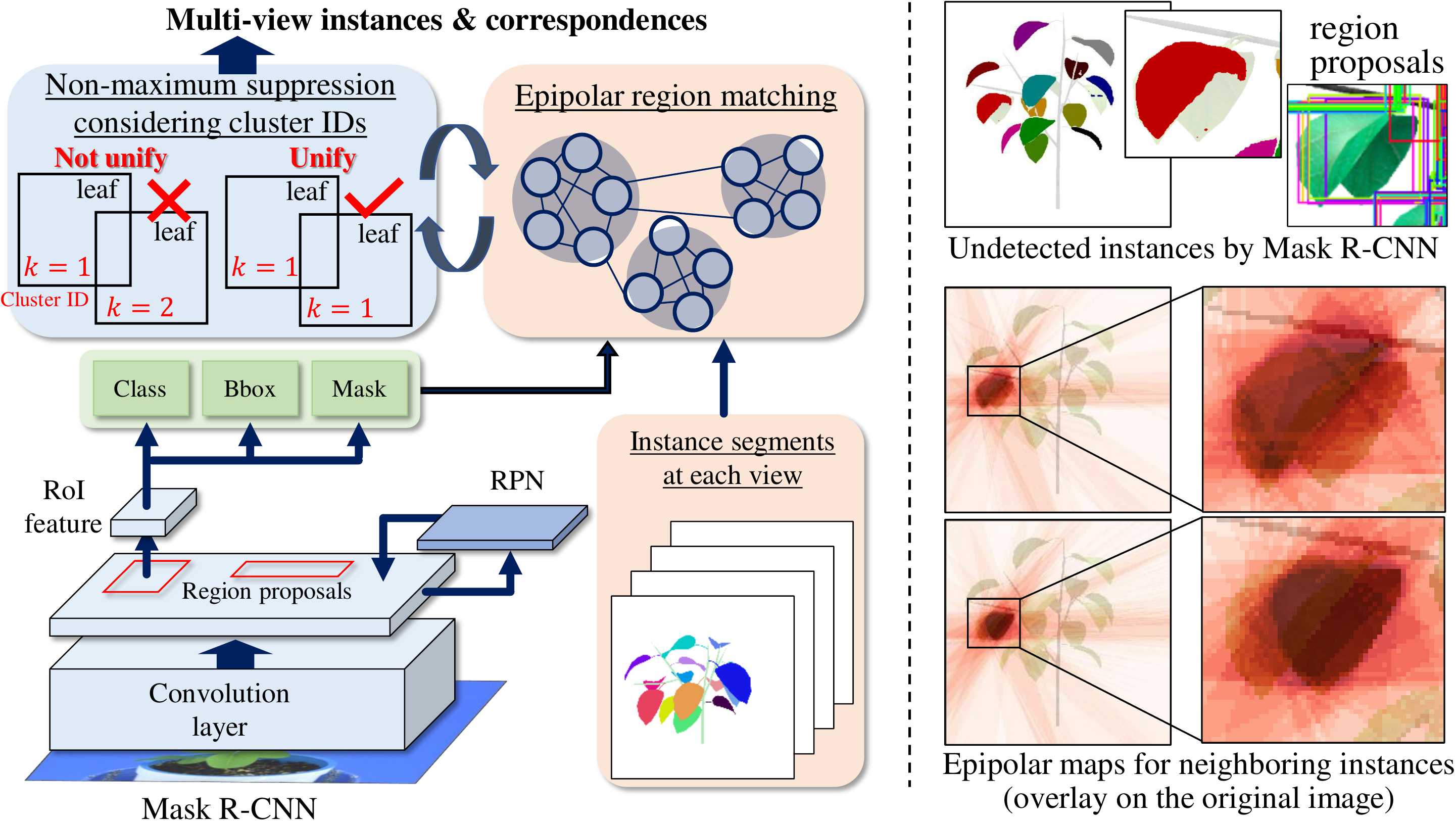}
\end{center}
   \caption{Epipolar region matching integrated in a two-stage instance segmentation network. It updates the instance detection and multi-view correspondence via an iterative process. The projections of epipolar bands are used to distinguish the undetected region proposal from neighboring instances.}
\label{fig:mvis}
\label{fig:roi_selection}
\end{figure}

Our method starts from the initial segmentation result, \eg, by Mask R-CNN, and their tentative multi-view correspondence computed by the epipolar region matching. 
The NMS and region matching processes can be performed alternately to update the set of detected instances and their multi-view correspondences as illustrated in \fref{fig:mvis}. 
We assign the instance cluster IDs for each region proposal by the region matching method, and avoid to unify the proposals with different cluster IDs by the NMS process.

To update the cluster ID of each region proposals, sets of corresponding instances (\ie, instance clusters estimated in the previous iteration) are projected to the view using the same manner as the epipolar band projection described in \sref{sec:epipolar}.
Let the $k$-th instance cluster in the previous iteration be $\{\region_k\}\subset\{\region\}$, which is denoted as a set of instances. 
The projection onto the $i$-th image, $\epimap_{i,k}$, is calculated based on the sum of epipolar bands by the instances (we call $\epimap_{i,k}$ an \emph{epipolar map} hereafter).
\begin{eqnarray}
    \epimap_{i,k} &=& \frac{\sum_{\{\region_k\}} \band(\region\in\{\region_k\},I_i)}{\#_c\{\region_k\}},
\end{eqnarray}
where $\#_c\{\region_k\}$ counts the number of instances in the cluster $\{\region_k\}$. The right side of \fref{fig:roi_selection} shows an example of the epipolar map.

An updated cluster ID for each region proposal is calculated based on the highest degree of intersection between the epipolar map and the region proposal.
Letting the instance mask of $r$-th region proposal in $i$-th image as $\region'_{i,r}$, the similarity of the two maps, $\epimap_{i,m}$ and $\region'_{i,r}$, is obtained by the similar manner to the IoU computation. 
Since the epipolar map does not take the value of $\{0,1\}$, we use an extension of the IoU, which is called Ruzicka similarity~\cite{deza2009encyclopedia}.
\begin{equation}
    s(\epimap_{i,k},\region'_{i,r}) = \frac{\Sigma_\vx \min(\epimap_{i,k}(\vx),\region'_{i,r}(\vx))}{\Sigma_\vx \max(\epimap_{i,k}(\vx),\region'_{i,r}(\vx))},
\end{equation}
where $\vx$ denotes the pixel location. In the equation, we deal with each instance mask $\region'_{i,r}$ as a map taking $\{0,1\}$ values, where the value takes one if the pixel is inside the mask. 
The updated cluster ID for the region proposal $\region'_{i,r}$ is selected as $k$ with the largest similarity $s(\epimap_{i,k},\region'_{i,r})$.
The update process of cluster ID is followed by the NMS that does not unify the region proposals with different cluster IDs to retain the partly occluded instances.
Our implementation allows to iterate the region matching and NMS processes while updating the set of instances and multi-view correspondences.

\subsection{Application: Instance-wise 3D reconstruction}
\label{sec:3d}

Multi-view correspondences of instance segments can be used for the {\it instance-wise} 3D reconstruction, by independently applying a 3D reconstruction method for each of instance clusters.
We implemented a simple volumetric reconstruction method based on a back-projection used in a traditional computed tomography~\cite{brooks75}, which is analogous to the visual hull method~\cite{laurentini1994visual}.

Let a set of $k$-th instance cluster $\{\region_k\}$ and the set of projection functions $\{\theta\}$ from 3D to 2D image coordinates, $\theta: \mathbb{R}^3 \rightarrow \mathbb{R}^2$.
The aggregated value $\mathcal{I}_{k}$ at voxel $\vx_{3D} \in \mathbb{R}^3$ can be computed as:
\begin{align}
    \mathcal{I}_k(\vx_{3D}) = \frac{\sum_{\region_k\in\{\region_k\}}\mathcal{L}_{\region_k}(\theta_k(\vx_{3D}))}{\#_c\{\region_k\}},
\label{eq:aggregation}
\end{align}
in which $\theta_k$ represents a projection from the voxel to the image coordinates corresponding to the instance $\region_k$.
Here, $\mathcal{L}_{\region_k}$ denotes a mask representation of $\region_k$, which returns $1$ if the pixel is inside the instance region $\region_k$.
The resultant voxel space $\mathcal{I}_{k}$ represents the ratio of voted instances; while we simply yielded the binarized version of the voxel space for the evaluation of the reconstruction accuracy, using the threshold of $\mathcal{I}_{k}(\vx_{3D})=0.5$.

\subsection{Implementation details}

\subsubsection{Region matching.}
During the epipolar band projection, we randomly sampled $200$ points in each instance to draw the epipolar lines.
The graph clustering is based on a Python implementation of SymNMF~\cite{kuang2012symmetric}.
Since the clustering algorithm requires the number of instance clusters $|k|$ as an input, we implemented a framework to search the optimal number of clusters. 
Assuming the instances are evenly occluded and are projected onto the similar number of views, 
we selected the optimal $|k|_{opt}$ with the minimum standard deviation of the number of instances contained in the clusters.
\begin{align}
    |k|_{opt} = \argmin_{|k|}\sqrt{\sum_{\region_k}(\#_c\{\region_k\}_{mean} - \#_c\{\region_k\})^2},
\label{eq:k}
\end{align}
where $\#_c\{\region_k\}_{mean}$ denotes the mean number of instances in instance clusters.
In the supplementary material, we provide a detailed analysis when the number of instance clusters $|k|$ is given (\ie, using the ground-truth number of objects).

\subsubsection{MVIS application.}
For the integration of the region matching for region proposals, we used a Keras implementation\footnote{\url{https://github.com/matterport/Mask_RCNN}} of Mask R-CNN.
We implemented an NMS using \texttt{numpy} and \texttt{nms} package outside the computation graph of Mask R-CNN.
To obtain the initial instances, we used the original Mask R-CNN with NMS by a large RoI threshold ($0.7$ in our experiment) and obtained excessive numbers of object RoIs with their instance masks.
During the iterations, our NMS is performed for each instance ID independently with a smaller threshold ($0.3$ was used in the experiment). 
With our unoptimized implementation, the whole process took up to several hours on a CPU (2.1 GHz, 8 threads); the projection of epipolar bands spent most of the time, which should be greatly optimized through a better implementation.

\section{Experiments}

We conducted experiments to assess the quality of the multi-view matching and instance-wise 3D reconstruction, which are core part of our framework.
%Simulated and real plant images were used for the quantitative evaluation, while we also introduce the results for other real-world scenes.
The supplementary material provides the detailed analysis and discussions, including the effect on the instance detection by our implementation.

\begin{figure}[tp]
 \begin{minipage}{0.5\hsize}
  \begin{center}
    \includegraphics[width=58mm]{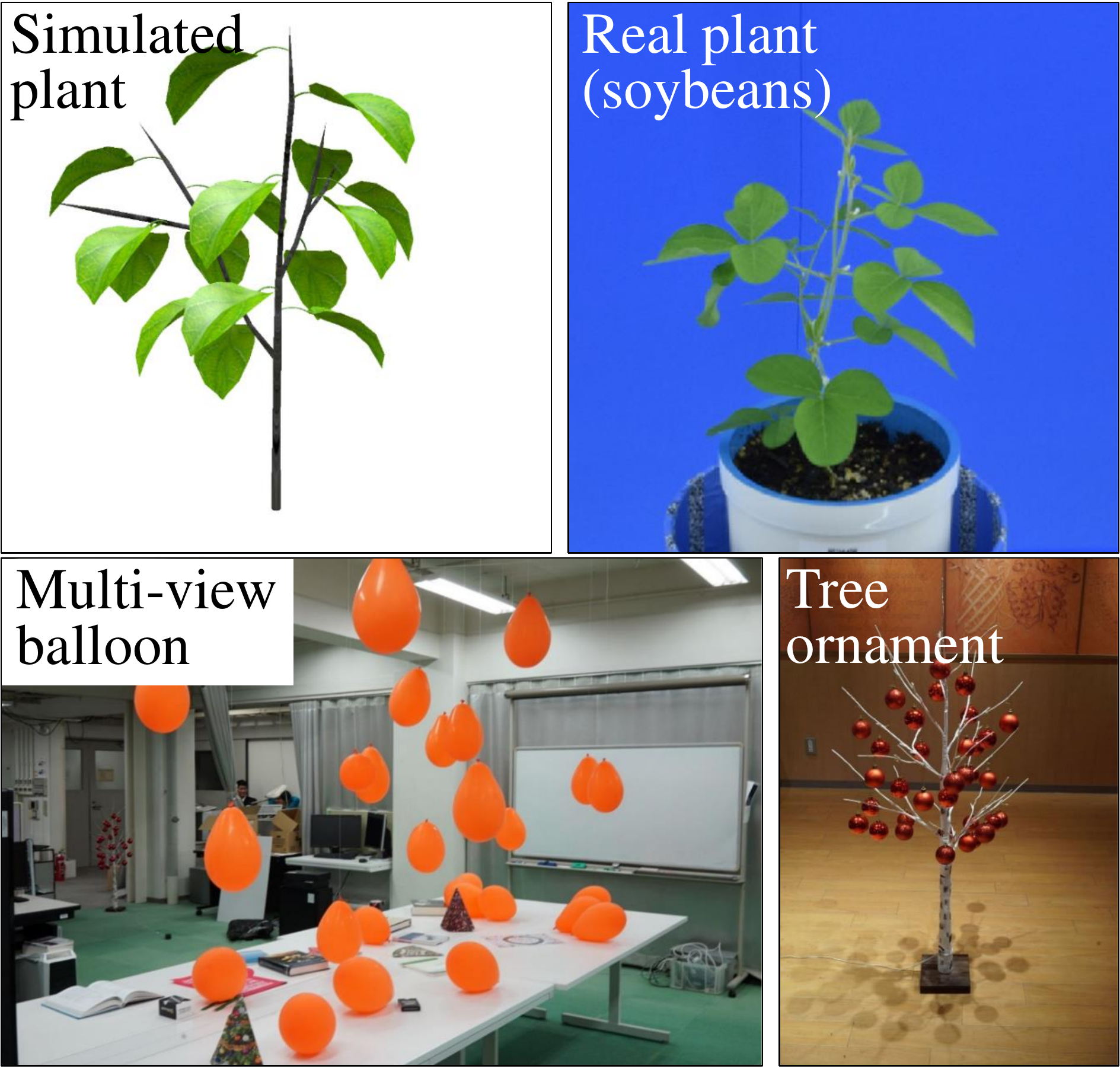}
  \end{center}
   \caption{Datasets used in the experiment.}
\label{fig:dataset}
 \end{minipage}
 \hspace{2mm}
 \begin{minipage}{0.45\hsize}
\begin{center}
    \includegraphics[width=51mm]{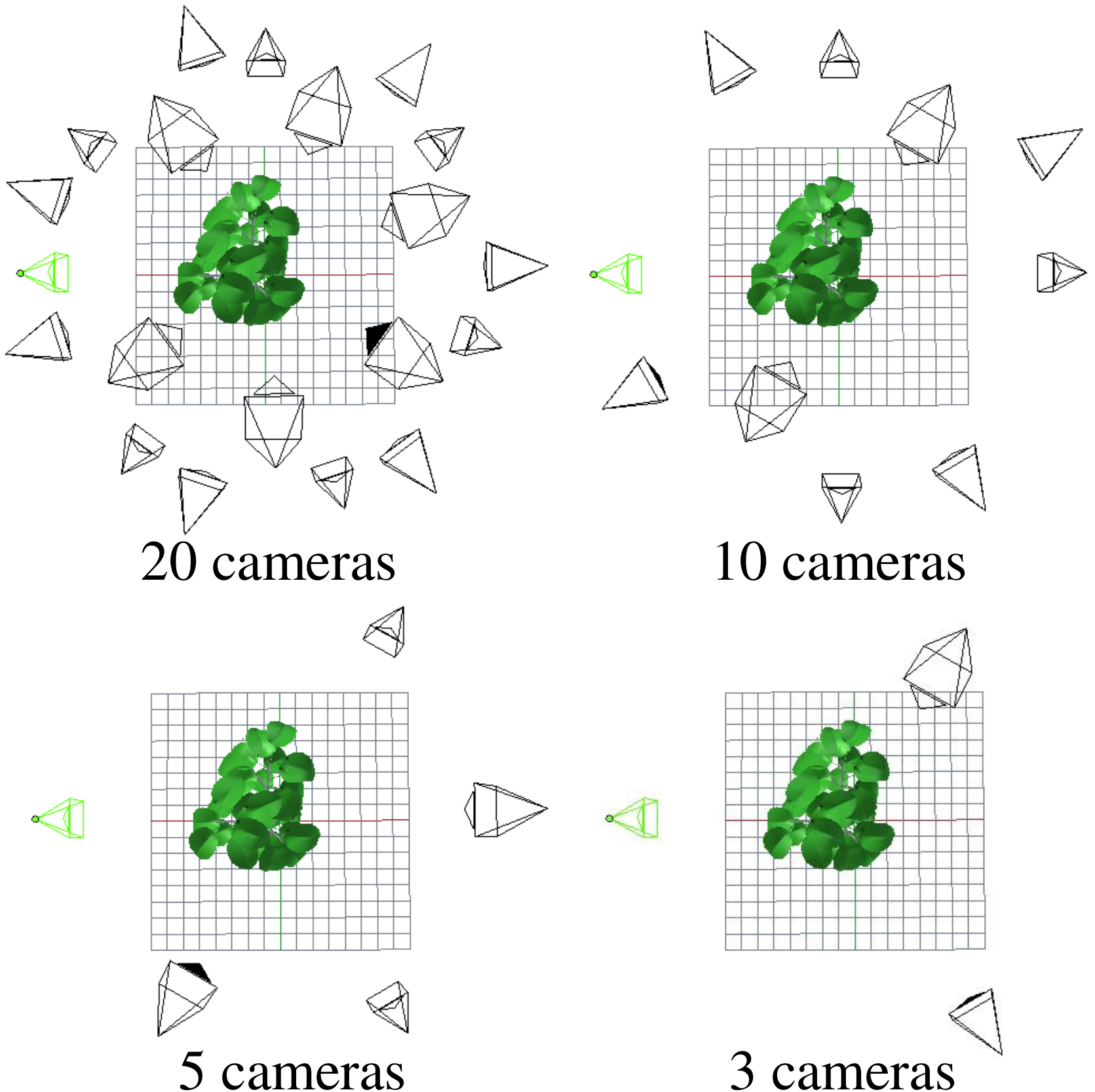}
\end{center}
   \caption{Camera setting for simulated plant dataset.}
\label{fig:view}
 \end{minipage}
\end{figure}

\subsection{Dataset}

We used the following datasets for the experiment (samples are shown in \fref{fig:dataset}). 

\vspace{-1em}
\subsubsection{Simulated plant dataset.}
We used simulated plant models, which were modified from a dataset used in a plant modeling study~\cite{CVPR18_plant}.
This dataset contains the ground truth instance masks and their multi-view correspondences, as well as the 3D shape information.
%The plant models were created by changing branching patterns and leaf positions using a self-organizing tree model~\cite{palubicki09}.
For training of Mask R-CNN, we prepared $14$ simulated plant models rendered from $32,760$ viewpoints ($458,640$ images in total).
For evaluation, we used four plant models with a different number of leaves ($4$, $8$, $16$, and $32$), not used for the training. 
To assess the effect of the varying number of views, we rendered the plants from different number of viewpoints ($3$, $5$, $10$, and $20$) illustrated in \fref{fig:view}, where we used the ground truth camera parameters.% for the experiment.

\vspace{-1em}
\subsubsection{Real-world dataset.}
We prepared the real-world scenes for experiment, in which the ground truth instance masks and multi-view correspondences were manually created.
The camera poses for these scenes were estimated via SfM~\cite{schoenberger2016sfm}.

\vspace{-0.5em}
\paragraph{Real plant (soybeans) scene} is a set of multi-view images of a soybean plant captured by a multi-view capturing system~\cite{tanabata2018development}, which was originally created for plant science studies.
We used $28$ views for instance matching. The number of distinctive leaves in the plant was $28$. 
COCO-trained Mask R-CNN was fine-tuned using the images of eight soybeans plants captured by the same system, which were not used during the evaluation.

\vspace{-0.5em}
\paragraph{Multi-view balloon scene} contains a scene with $27$ balloons captured from $18$ views. %. for the instance matching.
%As potential applications of MVIS, we performed the evaluations other than plant leaves.
%Inspired by the balloon dataset\footnote{\url{https://github.com/matterport/Mask_RCNN/tree/master/samples/balloon}} for instance segmentation, we captured a scene with $27$ balloons from $18$ views for the instance matching.  % 15 in the sky + 12 on the ground
Mask R-CNN, pre-trained with the COCO dataset, was fine-tuned using $61$ balloon images in the balloon dataset\footnote{\url{https://github.com/matterport/Mask_RCNN/tree/master/samples/balloon}}.

\vspace{-0.5em}
\paragraph{Tree ornament scene} contains an artificial tree with $33$ decorations captured from $21$ views.
%In a similar manner to the multi-view balloon dataset, we also experimented with a scene with an artificial tree with $33$ decorations captured from $21$ views.
Because the ornaments have spherical shapes, we fine-tuned the balloon-trained Mask R-CNN using a small number of (\ie, $25$) training images with tree ornaments, which were not used in the evaluation.

\subsection{Multi-view matching results}
% main result: DB_CG\MVS_result_summary

We evaluated the accuracy of multi-view region matching using the dataset.%simulated and real-world dataset. 

\vspace{-1em}
\subsubsection{Baselines.}
Because of no established baseline methods for descriptor-free region matching, we compared the proposed approach with straightforward implementations of instance matching 1) using MVS-based point cloud as guidance, 2) using the center point of regions, and 3) using sparse correspondences.

\vspace{-0.5em}
\paragraph{MVS-based matching.}
3D model by traditional MVS should be a guide for multi-view matching. 
We implemented a baseline matching method using dense 3D point clouds, called \emph{MVS-based matching} hereafter.
In this implementation, we used dense 3D point clouds reconstructed by COLMAP~\cite{schoenberger2016mvs}.
For $\region_{i,m}$, the $m$-th instance in the $i$-th view, we selected a 3D position $\vp_{\region_{i,m}}$ that represents the region of $\region_{i,m}$. 
An straightforward way to select $\vp_{\region_{i,m}}$ that is closest to the camera from among the points projected in the region of $\region_{i,m}$.
To increase the robustness to the outliers, we designated $\vp_{\region_{i,m}}$ as the centroid of the set of top $10\%$ closest points among those projected in the region.

We then project $\vp_{\region_{i,m}}$ to the other views; for example, if $\vp_{\region_{i,m}}$ is projected on the $n$-th instance in the $j$-th view, $\region_{j,n}$, we deemed $\region_{i,m}$ and $\region_{j,n}$ are corresponding instances.
In the same manner to our proposed method, we constructed a matching graph with $|V|$ nodes, where the edge weights record the correspondences; \eg, the edge between $\region_{i,m}$ and $\region_{j,n}$ are weighted by one. We solved the graph clustering using SymNMF~\cite{kuang2012symmetric}.
This baseline method expects that a good-quality 3D model is given, which is under the same assumption as the methods using point clouds as an input (\eg,~\cite{quan06}).

\vspace{-0.5em}
\paragraph{Point-based matching.}
Another baseline of the proposed approach is to match the centroid of instance regions instead of using the region matching. 
This method is analogous to the traditional \emph{point-based} epipolar matching, \ie, methods using epipolar \emph{lines} instead of \emph{bands}.
In our implementation, an epipolar line is drawn for the centroid of each instance region. 
The correspondences between instances in different views can be yielded as the nearest epipolar lines in 3D space. Similar to the MVS-based method, we constructed a matching graph, where the edge weights between the corresponding instances are weighted by one, and solved the graph clustering by SymNMF.

\vspace{-0.5em}
\paragraph{Matching based on sparse feature correspondences.}
We also assessed a multi-view region matching method using sparse correspondences, which is a typical example of feature descriptors.
We used AKAZE keypoints and descriptors~\cite{alcantarilla2011fast}. For each instance region, a corresponding instance for each different view is searched based on the number of corresponding points within the regions. 
We solved the graph matching similarly with the other baseline methods. This method is expected to work well when the scene is observed from similar viewing angles, which is a similar assumption to the video instance tracking.

\vspace{-1em}
\subsubsection{Evaluation metric.}
The matching accuracy $s_{match}$ was calculated by the number of correctly classified instances over the number of all instances $|V|$.
Because we solve the multi-view matching as a clustering problem, we need to associate estimated instance clusters and the ground-truth clusters.
We determined the ground-truth cluster ID $k_{gt}$ corresponding to the ID of an estimated cluster $k_{est}$, by searching the mode of $k_{gt}$ among instances belonging to $k_{est}$.
\begin{eqnarray}
    s_{match} = \frac {\sum_{k_{est}} \max_{k_{gt}\in\{k_{gt}\}} |\mathrm{id}(\region_{k_{est}}) \cap k_{gt}| }{|V|},
\end{eqnarray}
where $\mathrm{id}(\region_{k_{est}})$ denotes the estimated ID of the instance $\region_{k_{est}}$, while $\{k_{gt}\}$ is the set of ground-truth instance cluster ID.
Therefore, $|\mathrm{id}(\region_{k_{est}}) \cap k_{gt}|$ returns $1$ if the two instance IDs are same.
The number of {\it correct} matches is divided by $|V|$, the total number of instances among multi-view images.
As an evaluation of clustering problems, $s_{match}$ is equivalent to the {\it purity} metric, which is a common measure for evaluating the success of clustering. 

We also evaluated the accuracy of the estimated number of clusters $|k|_{opt}$ by computing the mean absolute error (MAE) $e_{|k|}$ of the estimated number of clusters via \eref{eq:k}.

%%%% 結果の表

\begin{table}[t]
%\small
\caption{Multi-view matching accuracy $s_{match}$ and MAE of the number of clusters $e_{|k|}$ for CG plant dataset. The values are averaged over the number of cameras and leaves. (Bold: the best accuracy among the methods.)}
\vspace{-1mm}
\begin{center}
\begin{tabular}{c|c| |c|c|c|c|| c|c|c|c|| c|}
\cline{2-11}
\multirow{2}{10mm}{\centering } &
\multirow{2}{18mm}{\centering Method}  & \multicolumn{4}{c||}{\# cameras}  & \multicolumn{4}{c||}{\# leaves} & \multirow{2}{13mm}{\centering Average} \\ \cline{3-10}
& & \multirow{1}{8.5mm}{\centering 20} & \multirow{1}{8.5mm}{\centering 10}	& \multirow{1}{8.5mm}{\centering 5}  & \multirow{1}{8.5mm}{\centering 3} & \multirow{1}{8.5mm}{\centering 32} & \multirow{1}{8.5mm}{\centering 16} & \multirow{1}{8.5mm}{\centering 8} & \multirow{1}{8.5mm}{\centering 4} &      \\ \hhline{~==========}
\multirow{4}{10mm}{\centering $s_{match}$\\ $\uparrow$}  & 
MVS-based     & {\bf 0.935}   & 0.735         & 0.211          & 0.215         &  0.337        & 0.584         & 0.543         &  0.631        & 0.524         \\ \cline{2-11}
&Point-based   & 0.715         & 0.697         & 0.521          & 0.463         &  0.477        & 0.497         & 0.612         &  0.810        & 0.599         \\ \cline{2-11}
&Sparse        & 0.392         & 0.488         & 0.562          & 0.509         &  0.376        & 0.455         & 0.500         &  0.620        & 0.488         \\ \cline{2-11}
&MVIS          & 0.905         & {\bf 0.910}   & {\bf 0.860}    & {\bf 0.856}   &  {\bf 0.655}  & {\bf 0.895}   & {\bf 0.987}   &  {\bf 0.994}  & {\bf 0.883}   \\ \hhline{~==========}

\multirow{4}{10mm}{\centering $e_{|k|}$\\ $\downarrow$}  & 
MVS-based      & {\bf 1.50}    & 7.25          & 13.25         & 9.25          &  17.00        & 10.00         & 3.25          &  1.00         & 7.81         \\ \cline{2-11}
&Point-based    & 19.50         & 25.75         & 4.00          & 8.50          &  11.75        & 16.75         & 20.25         &  9.00         & 14.44         \\ \cline{2-11}
&Sparse         & 31.00         & 26.00         & 21.50         & 9.25          &  15.25        & 25.75         & 22.25         &  24.50         & 21.94         \\ \cline{2-11}
&MVIS           & 2.00          & {\bf 2.25}   & {\bf 3.00}     & {\bf 2.50}      &  {\bf 9.00}   & {\bf 0.75}    & {\bf 0.00}   &  {\bf 0.00}  & {\bf 2.44}    \\ \cline{2-11}
\end{tabular}
\end{center}
\label{tab:matching}
\label{tab:k}

\vspace{-3mm}
\caption{Multi-view matching accuracy $s_{match}$ and MAE of number of clusters $e_{|k|}$ for real-world dataset. (Bold: the best accuracy among the methods.)}
\vspace{-1mm}
\begin{center}
\begin{tabular}{|c| |cc|cc|cc|| cc|}
\hline
\multirow{2}{20mm}{\centering Method} & \multicolumn{2}{c|}{Real plant}  &  \multicolumn{2}{c|}{Balloon}  & \multicolumn{2}{c||}{Ornament}  & \multicolumn{2}{c|}{Average} \\ \cline{2-9}
                    & \multirow{1}{12.8mm}{\centering $s_{match}\uparrow$} & \multirow{1}{10mm}{\centering $e_{|k|}\downarrow$}  
                    & \multirow{1}{12.8mm}{\centering $s_{match}\uparrow$} & \multirow{1}{10mm}{\centering $e_{|k|}\downarrow$}  
                    & \multirow{1}{12.8mm}{\centering $s_{match}\uparrow$} & \multirow{1}{10mm}{\centering $e_{|k|}\downarrow$}  
                    & \multirow{1}{12.8mm}{\centering $s_{match}\uparrow$} & \multirow{1}{10mm}{\centering $e_{|k|}\downarrow$}   \\ \hline\hline
%    method         &  $s_{match}$  &   $e_{|k|}$  &  $s_{match}$  &   $e_{|k|}$  &  $s_{match}$    &   $e_{|k|}$ & avg s       & avg e 
MVS-based           & 0.220         & 36             & 0.081        & 25          &  0.680          &  {\bf 1}    & 0.327       & 20.7          \\ \hline
Point-based   & 0.239         & 34             & 0.545        & 40          &  {\bf 0.683}    &  34         & 0.489       & 36.0          \\ \hline
Sparse        & 0.237         & 41             & 0.270        & 36          &  0.679    &  33         & 0.395       & 36.7          \\ \hline
MVIS      & {\bf 0.562}   & {\bf 26}       & {\bf 0.751}  & {\bf 3}     &  0.676          &  {\bf 1}    & {\bf 0.663} & {\bf 10.0}   \\ \hline
\end{tabular}
\end{center}
\label{tab:real}
\end{table}

%%%% 結果の表

\vspace{-1em}
\subsubsection{Result.}
\Tref{tab:matching} shows the the matching accuracy $s_{match}$ and the MAE of cluster numbers $e_{|k|}$ for the simulated plant dataset.
The table compares the accuracy and error averaged over different number of cameras and leaves.
The proposed MVIS implementation yielded a better accuracy for most cases and achieved the average matching accuracy of $88.3\%$, which outperforms the accuracy by the baseline methods.
%MVS-based matching ($52.4\%$), point-based matching ($59.9\%$), and the method based on sparse correspondences ($48.8\%$). 
Also, the result shows MVIS accurately estimates the number of clusters (\ie, the number of objects in the scene). %compared to the other methods.
MVS-based matching yielded better accuracy when using a larger number of cameras (\ie, $20$) because the smaller differences in the viewing angles enable it to find the dense texture-based correspondences.
In the cases using a smaller number of views, the MVS-based method notably drops the performance, although the proposed approach still achieved the matching accuracy of over $85\%$ when the number of views is $3$.

\Tref{tab:real} shows the matching performance for the real-world scenes. The proposed approach also achieved a reasonable accuracy for both multi-view matching and cluster number estimation. 
For the tree ornament scenes, the matching accuracy was comparable among the methods. Ornaments were small spheres, and well approximated as a point. 
Proposed MVIS has an advantage for the scenes with difficulties of dense 3D reconstruction (\eg, real plant scene) or with relatively large objects (\eg, balloon scene).

\subsection{Instance-wise 3D reconstruction results}

We here describe the quantitative result of 3D reconstruction using simulated plant models, which we have the ground-truth shape of the leaves. %Visual results of MVIS and instance-wise 3D reconstruction for the real-world dataset are also introduced in this section.

\vspace{-1em}
\subsubsection{Baselines.}
We used two baselines for the evaluation of 3D reconstruction accuracy, although these methods do not provide the instance-wise 3D reconstruction.
Since we used back-projection-based 3D reconstruction described in \sref{sec:3d}, we implemented a simple method using the back projection. 
Without relying on the instance correspondence, we unified the silhouettes of all leaves and inputted to the back projection.
This mimics the \emph{semantic segmentation} of leaves as the silhouette source, instead of using multi-view instances.
As another baseline, we simply used the \emph{MVS-based point clouds} reconstructed by COLMAP~\cite{schoenberger2016mvs}. 

\vspace{-1em}
\subsubsection{Evaluation metric.}
We evaluated the geometric error between the dense point clouds of the reconstructed and the ground truth leaf shapes.
For the evaluation, we first unify all 3D shapes of reconstructed leaf instances and convert the 3D voxel representation to a dense point cloud, where a 3D point is located if the voxel is inside leaves. 
The ground-truth leaf shape was originally modeled using polygons, we oversampled the vertices by Catmull–Clark subdivision~\cite{catmull1978recursively} to yield the dense point clouds.

Let $\vi \in \mathcal{I}$ and $\vt \in \mathcal{T}$ be estimated and the ground-truth 3D points, respectively. 
The geometric error is defined as a bidirectional Euclidean distance~\cite{zhu12} between the two point sets written as
\begin{align}
\small
d(\mathcal{I}, \mathcal{T}) = \frac{1}{2} \left( \frac{\sum_\mathcal{I} ||\vi - N_{\mathcal{T}} (\vi)|| }{|\mathcal{I}|} + \frac{ \sum_{\mathcal{T}} || \vt - N_{\mathcal{I}} (\vt)|| }{|\mathcal{T}|} \right), \nonumber
\end{align}
where $N_{\mathcal{I}}(\vx)$ and $N_{\mathcal{T}}(\vx)$ are functions to acquire the nearest neighbor point to $\vx$ from point sets $\mathcal{I}$ and $\mathcal{T}$, respectively, and $|\mathcal{I}|$ and $|\mathcal{T}|$ denote the numbers of points in $\mathcal{I}$ and $\mathcal{T}$.

\begin{figure}[t!]
\begin{center}
    \includegraphics[height=39mm]{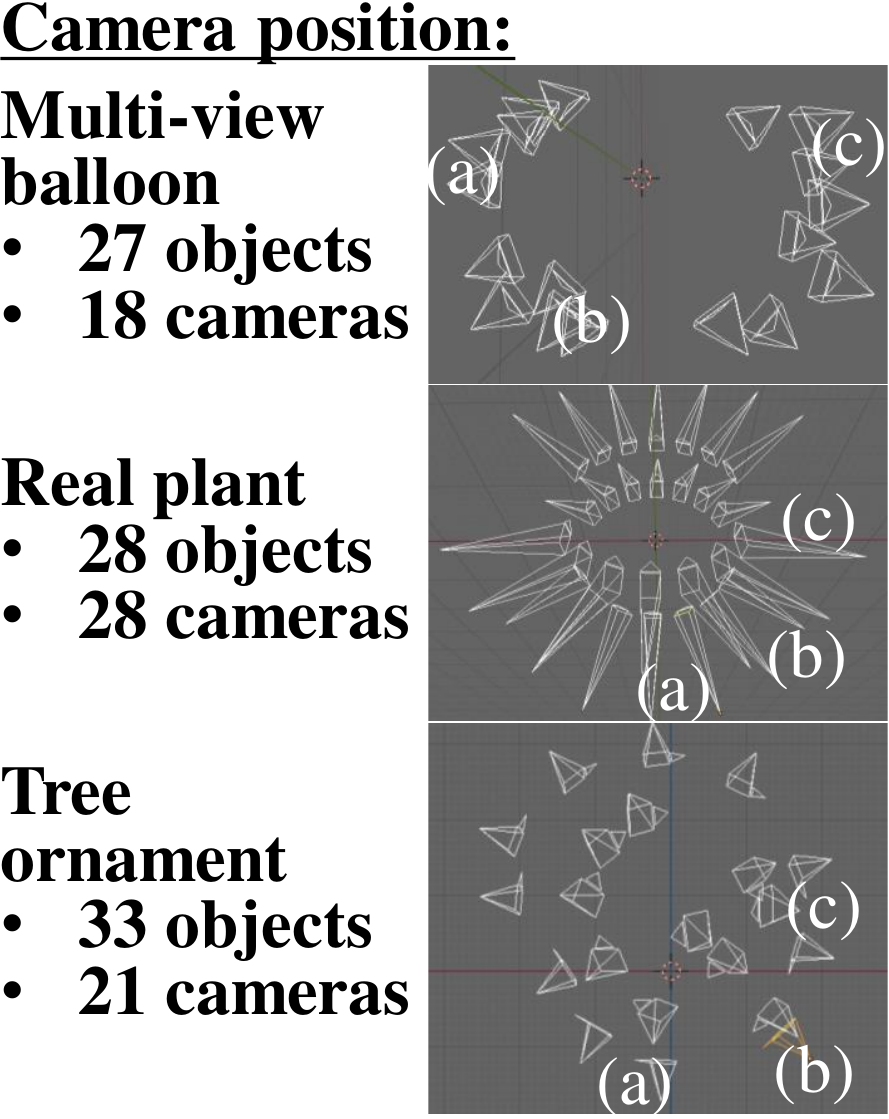}    \hspace{0mm}
    \includegraphics[height=39mm]{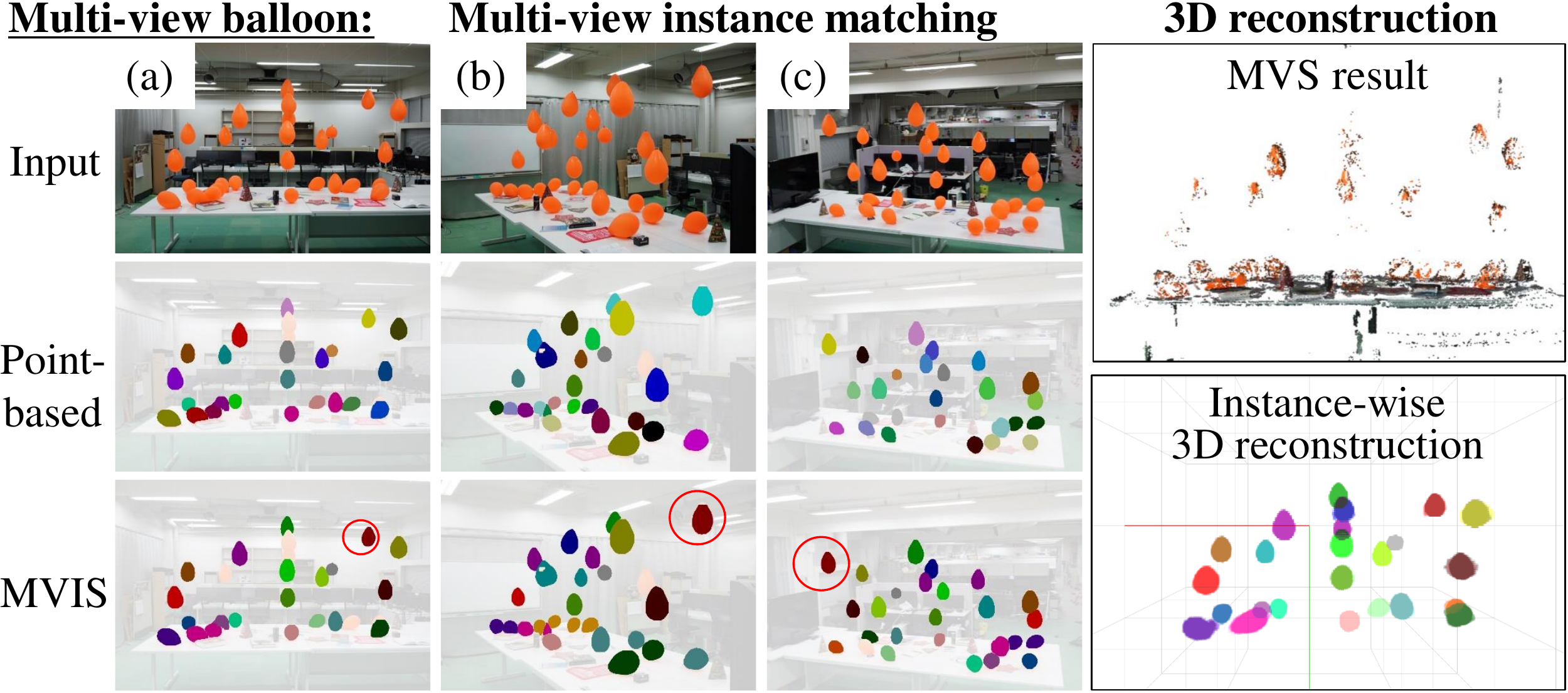} \\ \vspace{3mm}
    \includegraphics[height=35mm]{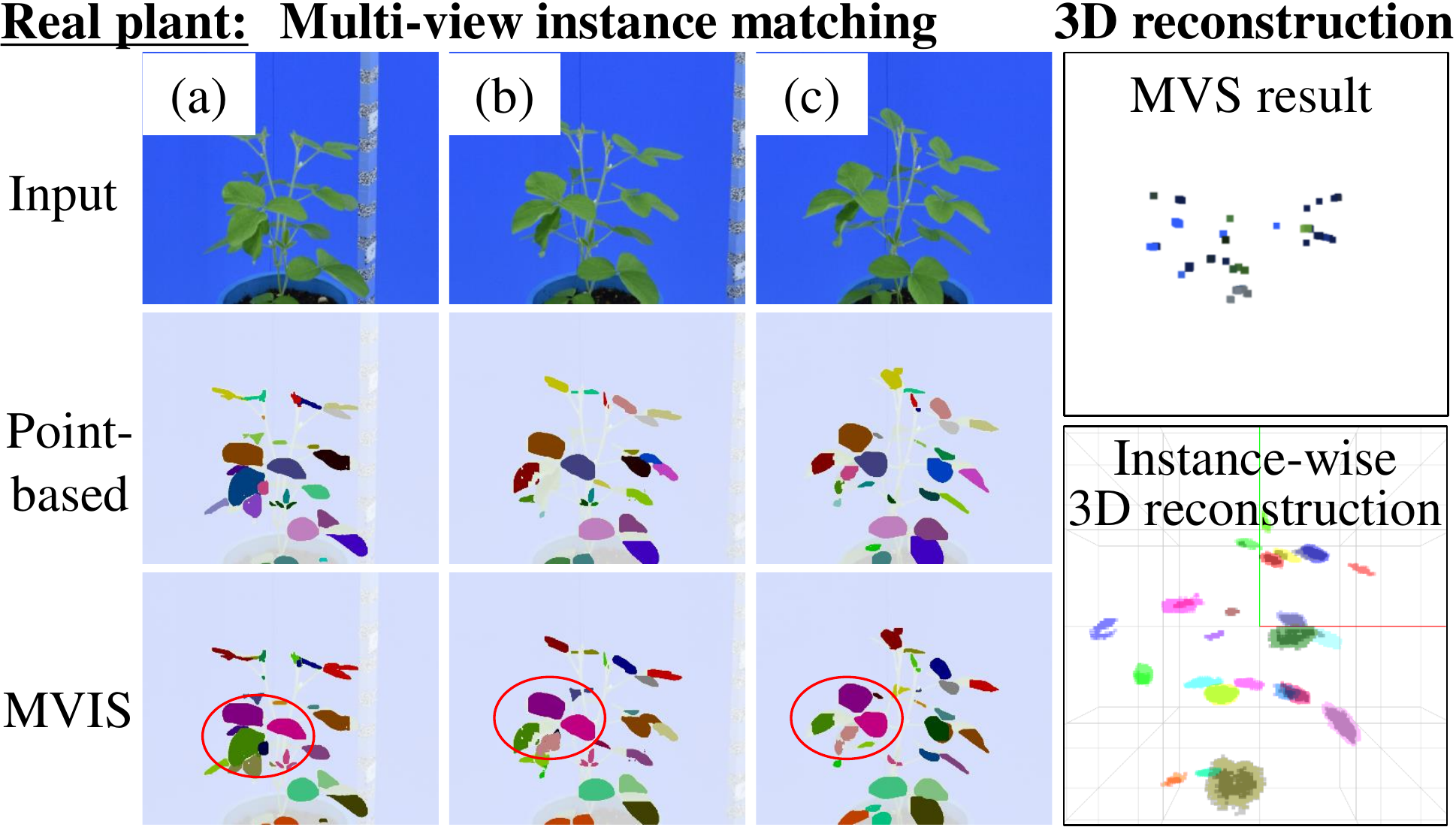} \hspace{-1mm}
    \includegraphics[height=35mm]{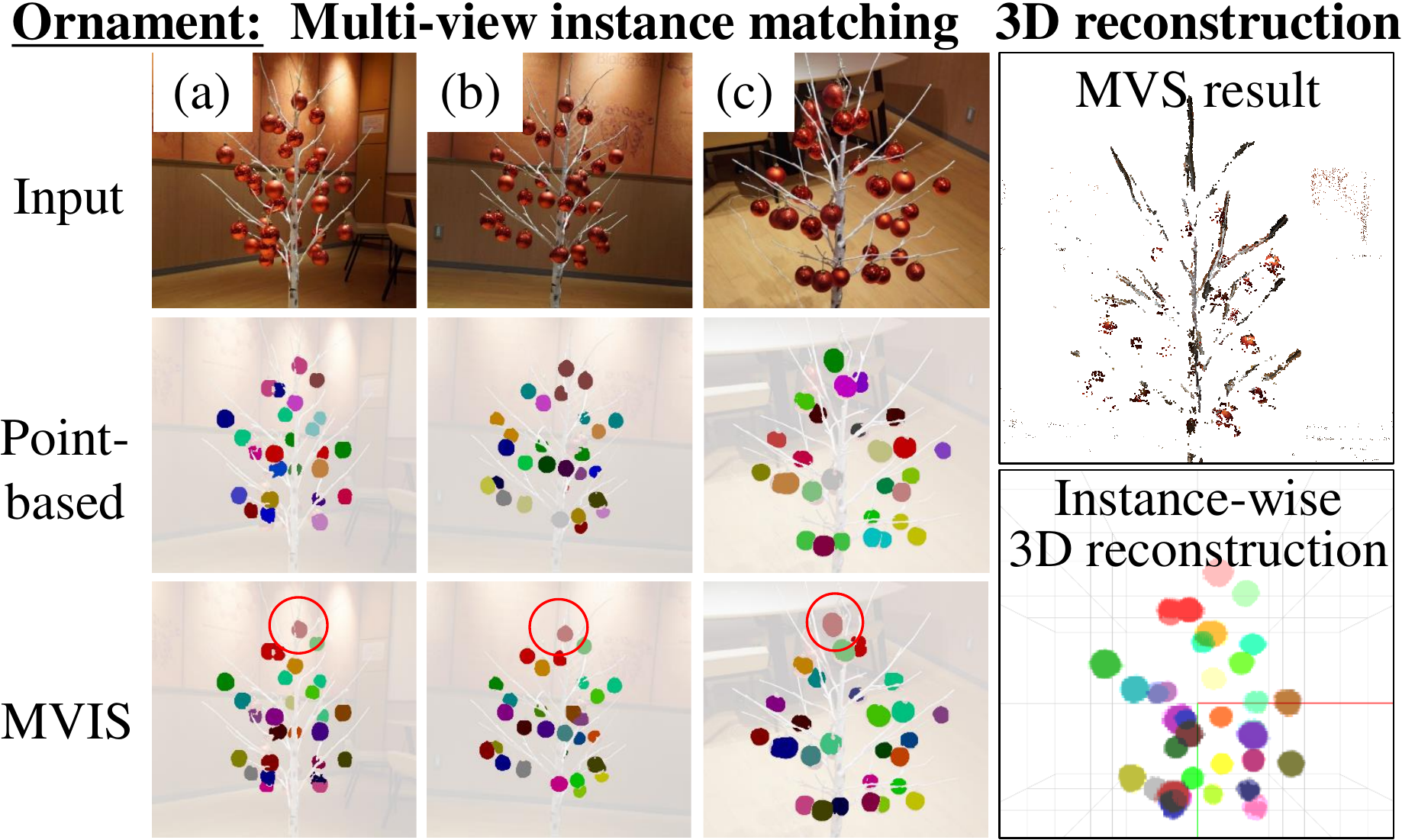}
\end{center}
   \caption{Visual results for real-world examples. 
   We show multi-view instances via the proposed MVIS and the point-based matching, which yielded a better matching accuracy than the other baselines.
   Although some instances are undetected or misclassified, MVIS yields convincing correspondences (\eg, the circled objects) and 3D reconstructions.}
\label{fig:visual}
\end{figure}

%%%%% TABLE: 3D reconstruction accuracy

\begin{table}[t]
%\small
\caption{3D reconstruction error $d(\mathcal{I}, \mathcal{T})$ normalized by the mean leaf length. N/A denotes the failure of the reconstruction (\ie, no point cloud was generated).}
\begin{center}
\begin{tabular}{|c| |c|c|c|c|| c|c|c|c|}
\hline
\multirow{2}{40mm}{\centering Method}  & \multicolumn{4}{c||}{\# cameras (averaged)}  & \multicolumn{4}{c|}{\# leaves (10 cameras) } \\ \cline{2-9}
                    & \multirow{1}{8.5mm}{\centering 20} & \multirow{1}{8.5mm}{\centering 10}	& \multirow{1}{8.5mm}{\centering 5}  & \multirow{1}{8.5mm}{\centering 3} & \multirow{1}{8.5mm}{\centering 32} & \multirow{1}{8.5mm}{\centering 16} & \multirow{1}{8.5mm}{\centering 8} & \multirow{1}{8.5mm}{\centering 4}     \\ \hline\hline
MVS (COLMAP)            & {\bf 0.027}   & 0.125         & N/A           & N/A           &  0.311        & 0.111         & {\bf 0.046}   &  {\bf 0.030}  \\ \hline
Semantic segmentation   & 1.593         & 1.707         & 2.556         & 2.068         &  1.306        & 2.269         & 1.396         &  1.858         \\ \hline
MVIS (proposed)         & 0.077         & {\bf 0.082}   & {\bf 0.185}   & {\bf 0.129}   &  {\bf 0.183}  & {\bf 0.057}   & {\bf 0.046}   &  0.042   \\ \hline
\end{tabular}
\end{center}
\label{tab:3d}
\end{table}

%%%%% TABLE: 3D reconstruction accuracy

\vspace{-1em}
\subsubsection{Results.}
\Tref{tab:3d} shows the 3D reconstruction error $d(\mathcal{I}, \mathcal{T})$. Since the geometric error is defined only up to scale like most multi-view 3D reconstruction methods, the errors were normalized by the average leaf length. 
For the comparison across the different number of cameras (left side of the table), we averaged the error over the different number of leaves. 
For the right half of the table, results using $10$ cameras are listed because MVS often failed the dense reconstruction for the smaller number of views.
The proposed method (MVIS) achieved better accuracy in most cases. 
Although the traditional MVS yielded an accurate reconstruction when using a larger number of (\ie, $20$) cameras or smaller number of leaves, the reconstruction was inaccurate or failed due to the difficulties of finding dense correspondences. 
%for large number of leaves (\eg, $32$) and failed the reconstruction using a few ($< 5$) cameras, due to the difficulties of finding dense correspondences. 
The average reconstruction error by the proposed MVIS did not notably drop when decreasing the number of views, which still achieved the error of $12.9~\%$ of the average leaf length via the reconstruction using $3$ cameras.

\vspace{-1em}
\subsubsection{Visual examples.}
\Fref{fig:visual} shows example results of MVIS and instance-wise 3D reconstruction for real-world datasets. 
In the MVIS result, corresponding instances are visualized with the same color. 
The proposed method yields the multi-view correspondences and instance-wise 3D shapes convincingly, although we do not have access to the ground-truth 3D shapes for real-world datasets.

\begin{wrapfigure}[8]{r}[0mm]{40mm}
\centering
    \vspace{-8mm}
    \includegraphics[keepaspectratio,width=40mm]{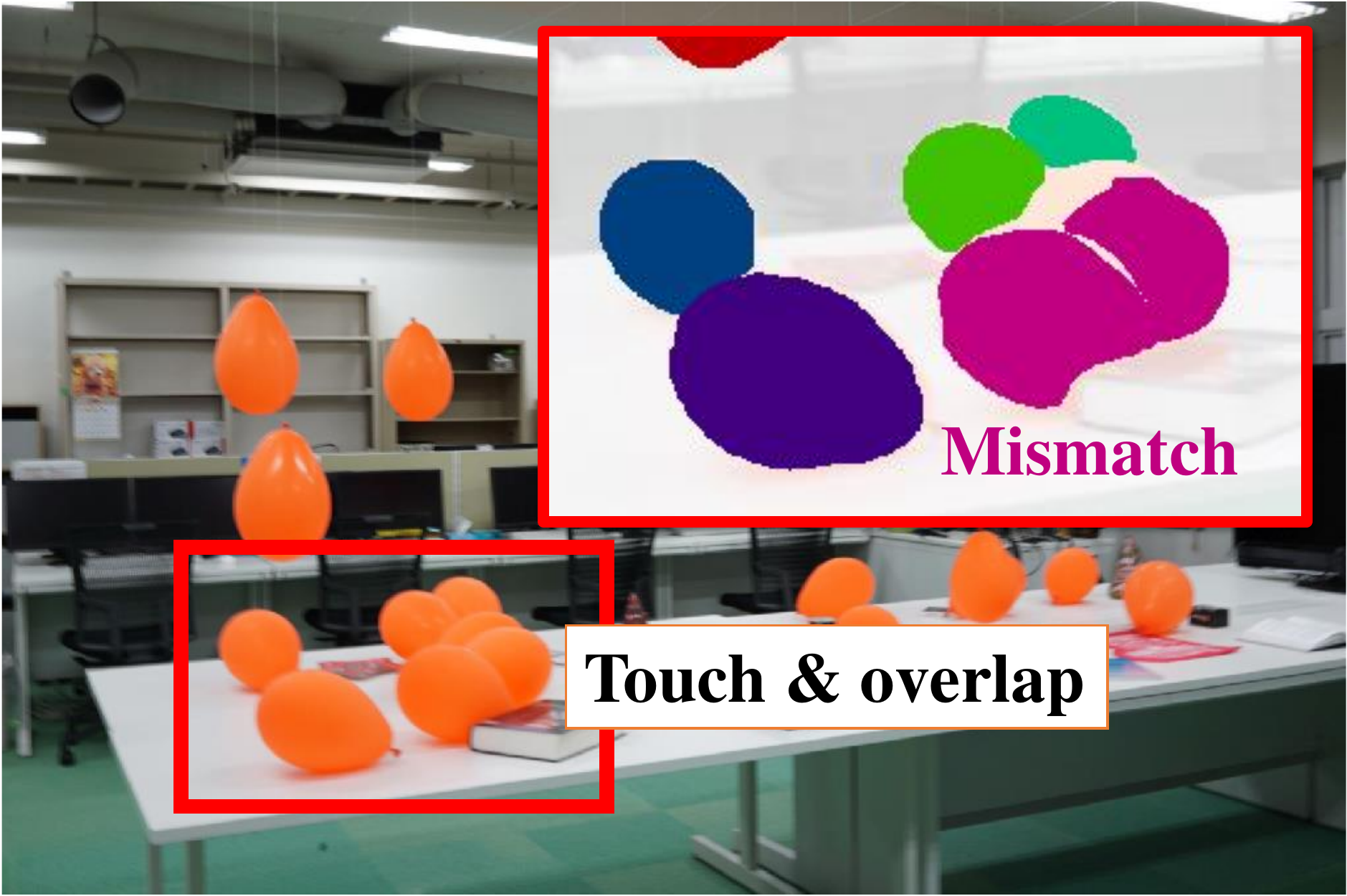}
    \caption{Failure case.}
\label{fig:touch}
\end{wrapfigure}

\vspace{-1em}
\subsubsection{Failure cases.}
Because the proposed method relies on the region segmentation, the failure in segmentation due to \eg, contact/occlusions of objects, affects the matching accuracy. Our experiment includes such cases when increasing the number of objects (\eg, \fref{fig:touch}). The low quality of instance masks is the dominant cause of the mismatching, which is a limitation of the proposed method.

\section{Discussion}
We introduced a multi-view matching method of object instances, which does not rely on the texture or shape descriptors instead of using geometric (\ie, epipolar) constraint and a graph clustering. 
Experiments with simulated plant models demonstrated the proposed method yielded the average accuracy of the multi-view instance matching over $85\%$, which outperforms the performances of baseline methods based on descriptor-based approaches such as MVS.
Our method also showed the potential to be used for instance-wise 3D reconstruction via the integration with 3D reconstruction methods such as the back projection.

Beyond the computer vision study, potential applications of the proposed method include the growth analysis of plants, as our experiments used a dataset from plant science and agricultural research field.

\vspace{2mm}
\noindent{\bf Acknowledgements.} This work was supported in part by JST PRESTO Grant Number JPMJPR17O3.

%===========================================================
\bibliographystyle{splncs}
\bibliography{cvpr2020doi}

%this would normally be the end of your paper, but you may also have an appendix
%within the given limit of number of pages
\end{document}